%% file: main.tex
\documentclass{template}
\usepackage{ragged2e}
\justifying

\input{header.tex}

\input{commands.tex}
\graphicspath{{media/}}

\usepackage{lineno}

\usepackage{hyperref}%

\begin{document}

\DeclareDelimFormat{finalnamedelim}{\printdelim{multinamedelim}}
\pagestyle{fancy}
\fancyhead[L]{}

\title{\sname{}: a fingertip-sized vision-based tactile sensor for robotic manipulation}

\maketitle

\author{Iris Andrussow*}
\author{Huanbo Sun}
\author{Katherine J. Kuchenbecker}
\author{Georg Martius*}

\begin{affiliations}

Corresponding authors' email addresses: \href{mailto:andrussow@is.mpg.de}{andrussow@is.mpg.de}, \href{mailto:georg.martius@tuebingen.mpg.de}{georg.martius@tuebingen.mpg.de}\\

I. Andrussow, K. J. Kuchenbecker: Haptic Intelligence Department, Max Planck Institute for Intelligent Systems, Heisenbergstr. 3, 70569 Stuttgart, Germany\\

I. Andrussow, H. Sun, G. Martius: 
Autonomous Learning Group, Max Planck Institute for Intelligent Systems, Max-Planck-Ring 4, 72076 Tübingen, Germany\\

\end{affiliations}


\setlength{\abovedisplayskip}{3pt}
\setlength{\belowdisplayskip}{3pt}

\thispagestyle{empty}

\begin{abstract}
Intelligent interaction with the physical world requires perceptual abilities beyond vision and hearing; vibrant tactile sensing is essential for autonomous robots to dexterously manipulate unfamiliar objects or safely contact humans. Therefore, robotic manipulators need high-resolution touch sensors that are compact, robust, inexpensive, and efficient. The soft vision-based haptic sensor presented herein is a miniaturized and optimized version of the previously published sensor Insight. Minsight has the size and shape of a human fingertip and uses machine learning methods to output high-resolution maps of 3D contact force vectors at 60 Hz. Experiments confirm its excellent sensing performance, with a mean absolute force error of 0.07 N and contact location error of 0.6 mm across its surface area. Minsight's utility is shown in two robotic tasks on a 3-DoF manipulator. First, closed-loop force control enables the robot to track the movements of a human finger based only on tactile data. Second, the informative value of the sensor output is shown by detecting whether a hard lump is embedded within a soft elastomer with an accuracy of 98\%. These findings indicate that Minsight can give robots the detailed fingertip touch sensing needed for dexterous manipulation and physical human–robot interaction.
\end{abstract}

\section{Introduction}

Autonomous robots have the potential to become dexterous \cite{OpenAI,Nagabandi2019:DexManipul,Quadrupedal} and work flexibly together with humans \cite{screw_unfastening, I_support}. 
To achieve this goal, their hardware needs to become more robust and provide richer sensory feedback, while their learning algorithms need to become more data-efficient and safety-aware.  
A clear shortcoming of current commodity robotic hardware is the complete lack or low quality of the tactile sensations it can acquire.
In contrast, humans have a rich sense of touch and use it constantly -- mostly subconsciously. In fact, if haptic perception is impaired, dexterous manipulation becomes very challenging or even impossible \cite{Ciadra2019:no-haptic-human}.
High-resolution haptic sensing similar to the human fingertip can enable robots to execute delicate manipulation tasks like picking up small objects, inserting a key into a lock, or handing a full cup of coffee to a human.
Thus, improving haptic sensors for learning robots should be of high priority.

Indeed, many haptic sensors have recently been invented \cite{wang_tactile_2019, BioTac0,e-Skin, HexoSkin,Ferroelectric,piacenza_sensorized_2020,GelSight,GelTip,lepora_digitac_2022,GelForce,SlimFEM,Eletroluminescent, LeeCapacitive, OmniTact, YanHallEffect,Beads,Romero2020GelSight2.5, Hellebrekers,ResistiveTaunyazov-RSS-20,lambeta_digit_2020}, but they are still far from the capabilities of human skin. 
Many of these sensors are too fragile, have low fidelity, cannot measure shear forces, or have a very limited sensing area (see also \secref{sec:RelatedWork}).
Vision-based haptic sensors are an exciting emerging approach to tackle these issues.
This approach uses a camera to monitor changes in internal light intensity and/or color caused by deformations of the sensor's surrounding material due to external contact forces. Analytical or data-driven methods are used to map these visible changes to tactile contact information. The centrally positioned sensing component, a camera, does not bear the contact load, which ensures high durability of the sensor. Furthermore, advanced image-processing techniques provide inspiration for effective ways of converting image data into tactile contact information.
The recently proposed \emph{Insight} sensor \cite{Sun2022:Insight} has many favorable properties for use in robotic fingertips, as its robust and inexpensive design provides all-around sensation of normal and shear forces at high resolution.

\begin{figure*}[t]
    \centering
    \includegraphics[width=0.95\textwidth]{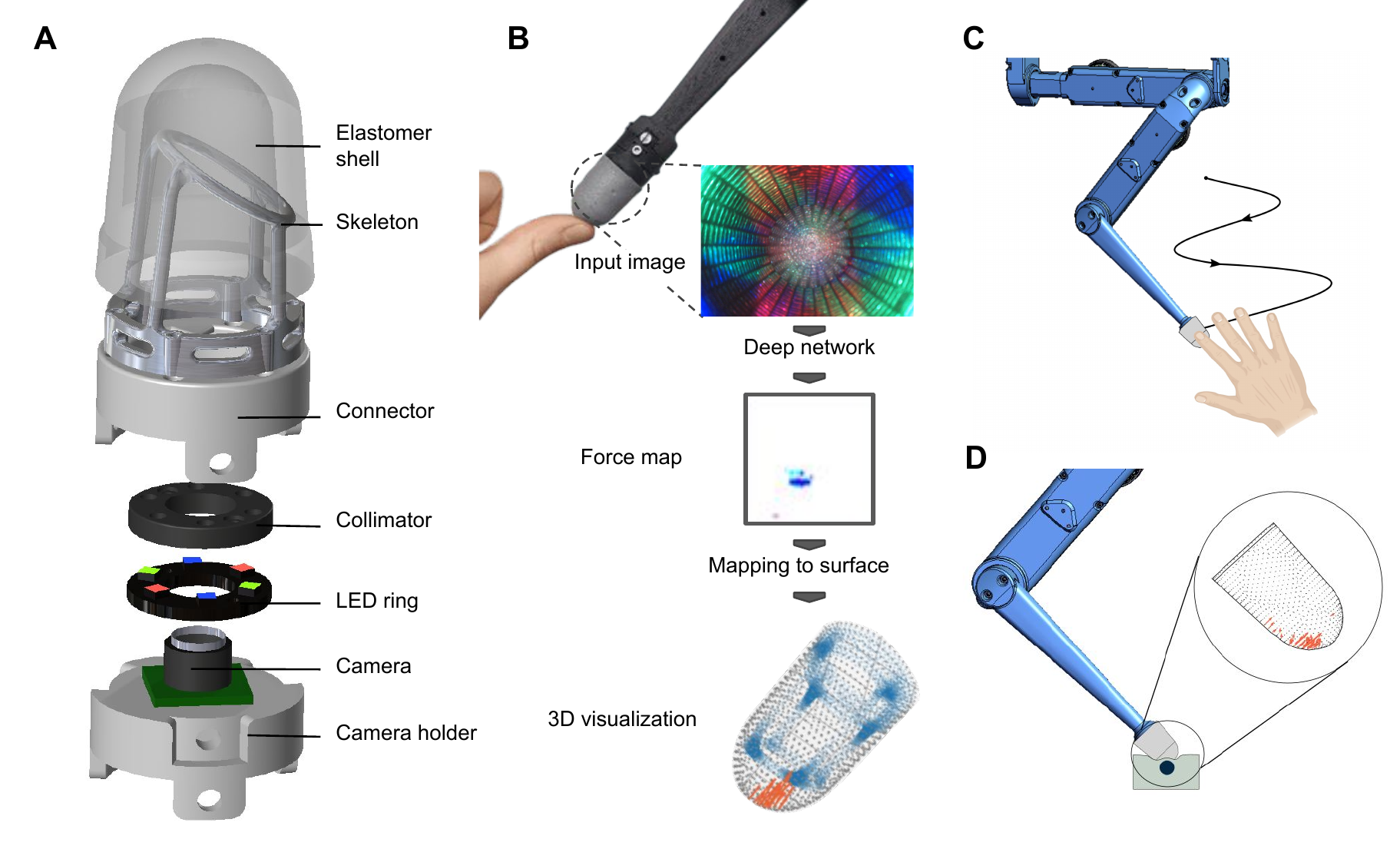}
    \caption{\sname{}'s design, processing pipeline, and conducted experiments. \textbf{A}: Exploded view of the proposed sensor. \textbf{B}: \sname{} mounted on the TriFinger robot platform and pressing into a human finger, along with the captured image, the sensor's processing pipeline, and the resulting set of force vectors. \textbf{C}: Tactile servoing experiment, where the robot maintains gentle contact with a moving human fingertip. \textbf{D}: Lump detection experiment, where the robot presses into simulated tissue and determines whether a hard lump is present.}
    \label{fig:overview}
\end{figure*}

However, as proposed in \cite{Sun2022:Insight}, Insight still has several shortcomings that we aim to eliminate in this work.
Namely, this sensor is too large (\SI{70}{mm} height, \SI{40}{mm} diameter) to fit on a typical robotic gripper or robot hand, and its CPU data-processing pipeline is too slow (\SI{10}{Hz}) to support real-time control.  
This paper proposes the design of a miniature sensor based on Insight's principles; \emph{\sname{}} reduces the sensor volume roughly by a factor of five, which means it is now comparable in size to the tip of a human thumb and can therefore be used with dexterous robot manipulators (\figref{fig:overview}). For particular real-time robotic applications, one needs to find a good balance between the hardware and software parts of the sensor design, such as suitable sensor geometry, adequate implementation speed, and sufficiently rich haptic output. We investigate the changes that are necessary to adapt Insight's design to \sname{}'s miniature size and realistic robotic applications. We thoroughly analyze the machine-learning pipeline that performs the force inference to obtain low computational complexity and an update rate that is six times faster (\SI{60}{Hz}). 
Besides rigorous quantitative evaluation, we demonstrate our new sensor's real-time capabilities in a tactile servoing task, where a robot arm follows a human fingertip just by feeling the contact forces. We furthermore leverage the inferred tactile data to detect hard lumps embedded in a soft elastomer, a clinically motivated task that requires accurate and distributed tactile sensing.

\section{Related Work}\label{sec:RelatedWork}
Numerous tactile sensors have been proposed in recent years to enhance the capabilities of robotic manipulators. One common approach is to arrange many small, discrete sensing elements in a grid and measure contact locally in a distributed way \cite{HexoSkin, LeeCapacitive,de_clercq_soft_2022,superresolution}.
Distributed haptic sensing that is based on resistive, capacitive, or barometric principles thus requires numerous sensing elements and wires that constantly suffer from bending and twisting loads, enduring a high risk of damage.
Along with problems like manufacturing complexity, this strategy becomes less favorable when high-resolution sensing is desired. 
Vision-based sensors are a promising alternative to achieve this goal. 

\tabref{tab:sensor_comparison} gives an overview of state-of-the-art tactile sensors that are intended for use on robotic fingers or manipulators. Many existing sensors are designed with a flat, uni-directional sensing surface and are targeted for use with conventional two-jaw parallel grippers, e.g., \cite{Beads,taylor_gelslim30_2021,bhirangi_reskin_2021,lambeta_digit_2020}. However, human fingers have omnidirectional sensing capabilities and feature a curved sensing surface, which is useful because curved fingertips do not need to reorient to maintain a perceivable contact when moving along an object \cite{Romero2020GelSight2.5}.

Despite the importance of fingertip convexity, only a few curved tactile sensors have been developed. The BioTac sensor \cite{lin_estimating_2013} uses multiple modalities (impedance of internal fluid measured by nineteen electrodes, a hydrophone, and a thermistor) to achieve high performance in contact-force estimation and vibration sensing. However, it is an expensive and delicate piece of hardware that therefore cannot be widely adopted in robotics.

Magnetic sensors \cite{bhirangi_reskin_2021,tomo_covering_2018} take advantage of the propagation properties of magnetic fields, which make it possible to place a grid of sensing elements below an elastomeric cover without any electric connections to the sensing surface. This feature can make them more robust than conventional capacitive or resistive approaches. However, the magnetic sensing approach is vulnerable to variations in object material, particularly ferromagnetic objects and nearby magnetic fields.

To achieve omnidirectional sensing, Epstein et al.~\cite{epstein_bi-modal_nodate} use machine-learning techniques to infer contact location and forces from eight barometric pressure sensors embedded within a rubber hemisphere, which makes the sensor robust to impacts. However, the sensed force range for this curved sensor starts only at \SI{0.5}{N}, whereas humans are capable of sensing much smaller contact forces~\cite{lederman_sensing_1999}. SaLoutos et al.~\cite{saloutos_design_2022} combine barometric pressure sensors and a neural network to estimate 3D contact forces and 2D contact locations. Their approach is robust and computationally fast, with a force sensing range up to \SI{25}{N}; the force error of \SI{1.58}{N}, however, also falls far short of human capabilities.  
Piacenza et al.~\cite{piacenza_sensorized_2020} use data-driven methods to extract contact location and normal forces from overlapping light signals guided within a transparent layer covering a fingertip-shaped sensor. Their sensor has a form factor far bigger than a human fingertip, and the sensed force range starts at \SI{0.2}{N}, which still leaves room for improvement.

\newcommand{\mrb}[2]{\multirow{#1}{*}{\rotatebox{90}{#2}}}
\newcommand{\ph}{\phantom{$^{b}$}}

\begin{table}[ht]
\centering
\caption{Comparison between Minsight and fourteen previously published small-scale tactile sensors that are intended for use on robot fingers. Given the importance of omnidirectional sensing, the selection of flat sensors has been limited to four salient examples. Unreported values are indicated by --.}
\resizebox{\textwidth}{!}{    
\begin{tabular}{@{}c@{}cp{.18\linewidth}@{}c@{\!\!}r@{}r@{\ \ \ }c@{\ \ \ }c@{}}
    \toprule
  \multicolumn{2}{c}{\multirow{2}{*}{Sensor}} & \multirow{2}{*}{Modality}                                                                    & W/$\varnothing \times$H & \multicolumn{1}{c}{Sensing} & \multicolumn{1}{c}{Sampling}  & Normal / Shear   & Localization \\
                                              &                                                 &                                            & [mm] & \multicolumn{1}{c}{Area [\si{\milli\metre\squared}]} & \multicolumn{1}{c}{Rate [Hz]}    & Force Error [N]  & Error [mm]   \\
\midrule
\mrb{4}{flat}                                              
                                              & ReSkin \cite{bhirangi_reskin_2021}              & magnetometers                              & 20$\times$20        & 400$^b$                              & 400\ph                            & 0.05 / 0.03      & 0.7         \\
                                              & Digit \cite{lambeta_digit_2020}                 & camera                                     & 19$\times$16        & 304\ph                                & 60$^c$                         & -- / --          & --           \\
                                              & GelSlim 3.0 \cite{taylor_gelslim30_2021}        & camera                                     & $\varnothing$28$^a$ & 675\ph                                & 90$^c$                         & -- / --          & --           \\
                                              & Sferrazza et al. \cite{Beads}\cite{Beads2}      & camera                                     & 32$\times$32        & 900\ph                              & 50\ph                             & 0.04 / --\hspace{3ex}        & --           \\
\midrule
\mrb{5}{curved}
                                              & BioTac \cite{lin_estimating_2013}               & electrical impedance, thermistor, hydrophone          & 15$\times$15        & 484\ph                                & \makecell[tr]{100$^d$\\ 2200$^e$} & 0.26 / 0.48        & 1.4          \\
                                              & uSkin \cite{tomo_covering_2018}                 & magnetometers                               & 35$\times$28        & $\sim$2000$^b$                       & 30\ph                             & 0.20 / 0.20        & 6/10         \\
                                              & Romero et al. \cite{Romero2020GelSight2.5}      & camera                                     & \phantom{0.}28$\times$30.5      & 2069\ph                               & 90$^c$                         & -- / --          & --           \\
\midrule
\mrb{9}{360$^\circ$}
                                              & Epstein et al. \cite{epstein_bi-modal_nodate}   & piezoresistors                             & $\varnothing$22$\times$11\ph    & 184\ph                                & 200\ph                            & 0.50 / --\hspace{3ex}         & 1.1$^b$          \\
                                              & SaLoutos et al. \cite{saloutos_design_2022}     & barometers,\newline time-of-flight sensors & $\varnothing$20$\times$10\ph    & 406\ph                                & 200\ph                            & 1.58 / --\hspace{3ex}         & 1.3$^b$          \\
                                              & Piacenza et al. \cite{piacenza_sensorized_2020} & photodiodes                                & $\varnothing$36$\times$72\ph    & 6107$^b$                            & 60\ph                             & 0.20 / --\hspace{3ex}         & 2.9          \\
                                              & TacTip \cite{ward-cherrier_tactip_2018}         & camera                                     & $\varnothing$40$\times$20\ph    & 2500$^b$                               & 30$^c$                         & -- / --          & 2.4          \\
                                              & GelTip \cite{gomes_geltip_2020}                 & camera                                     & $\varnothing$15$\times$35$^b$   & 2513$^b$                               & --\ph                             & -- / --          & 6.8          \\
                                              & OmniTact \cite{OmniTact}                        & camera                                     & $\varnothing$30$\times$33\ph    & 3110$^b$                             & 30$^c$                         & -- / --          & 2.3         \\
                                              & Insight \cite{Sun2022:Insight}                  & camera                                     & $\varnothing$40$\times$70\ph    & 4800\ph                               & 11\ph                             & 0.03 / 0.03      & 0.4          \\
                                              & \sname{} (ours)                                 & camera                                     & $\varnothing$22$\times$30\ph    & 1740\ph                               & 60\ph                             & 0.06 / 0.03      & 0.6          \\
  \bottomrule
\end{tabular}
}
{\footnotesize $^a$~GelSlim 3.0 has a round, flat sensing area with a diameter of $\sim$\SI{28}{mm}; 
$^b$~Estimation by the authors of this paper, based on other numbers reported in the original work; 
$^c$~Camera frame rate without any data processing; 
$^d$ Electrical impedance, temperature, heat flux, and DC pressure; 
$^e$~AC pressure.}
\label{tab:sensor_comparison}
\end{table}

In contrast to solutions with many small sensing elements, vision-based haptic sensors typically follow a centralized sensor arrangement strategy and use one or more cameras that view a soft contact surface from within \cite{GelTip,OmniTact}.
Advantages over solutions with many small sensing elements are that no extra wiring is required and that cameras can acquire high-fidelity data simultaneously across a broad spatial field at low cost. Many proposed vision-based sensors consist of multiple functional layers \cite{lepora_digitac_2022,GelForce}, which take time to fabricate properly and can tear apart. Other designs use multiple cameras \cite{OmniTact} and therefore have higher cost, complexity and size. The principle used for Insight \cite{Sun2022:Insight} and now \sname{} requires only a single camera and simple manufacturing procedures to infer contact locations and forces over the whole sensing surface. Outputting a distribution of force vectors provides detailed information about multiple contacts and 3D force directions, which is unique among current vision-based tactile sensors. Compared to the existing sensors listed in \tabref{tab:sensor_comparison}, \sname{} is the smallest sensor that accurately senses normal force, shear force, and contact location across a comparatively large 360$^\circ$ sensing surface. 

\section{The \sname{} Fingertip Sensor}
\label{sec:Design_Fabrication}

We adapt the working principle of the recently proposed \emph{Insight} sensor \cite{Sun2022:Insight} to create a miniature version -- \emph{\sname}, as presented in \figref{fig:overview} -- that delivers robust and dense information about contact forces and locations.  
It is a vision-based haptic sensor that infers distributed contact-force information from the camera image.
\sname{} is both mechanically and computationally optimized for the fingertips of robotic hands and manipulators like the TriFinger platform \cite{wuthrich_trifinger_2020}.

 \begin{figure*}
    \centering
    \includegraphics[width=1.0\textwidth]{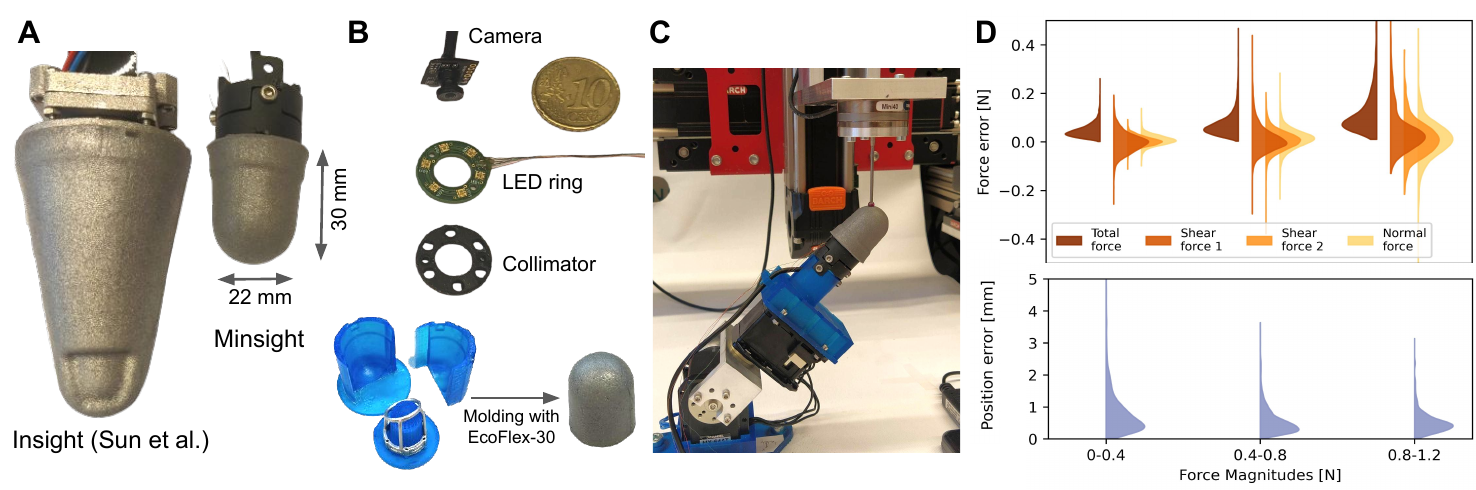}
    \caption{\textbf{A}: The new \sname{} sensor compared in size to the original Insight sensor. \textbf{B}: Top: USB camera and LED ring with collimator next to a European 10 cent coin (\SI{19.75}{mm} diameter). Bottom: We use a 3D-printed mold with an insert that holds the metal skeleton to mold the soft sensor shell with Smooth-On EcoFlex 00-30. \textbf{C}: The 5-DoF test bed includes an indenter mounted on a force-torque sensor to collect ground-truth force data for training the mapping. \textbf{D}: Distribution of force and position errors across different force magnitudes.}
    \label{fig:minsight_package}
\end{figure*}

As described in \cite{Sun2022:Insight}, the sensing principle relies on a camera that detects forces applied to the outside of a soft opaque elastomer shell by observing the propagated deformations on the inside. 
To increase the sampling rate and reduce the size, we replace the original Raspberry Pi camera with a small-scale USB 3.0 camera that captures \res{1280}{720} images at a frequency of \SI{60}{\hertz}; these widescreen images are cropped to the central \res{1020}{720} where the shell is visible. By eliminating the bottleneck of transferring images from the Raspberry Pi to the processing hardware, we can achieve an update rate that is six times faster than that of Insight~\cite{Sun2022:Insight}. The USB interface simplifies integration into the host platform and allows a direct connection to the compute hardware. The camera is located near the bottom of the skeleton and equipped with a \num{160}$^\circ$ fish-eye lens, so it can observe the entire sensitive part of the inside surface of the over-molded shell while keeping the overall form factor of the sensor compact.

The sensor uses a combination of structured light and shading effects to reconstruct the 3D deformations with its single camera.
The structured light pattern is created by a custom 3D-printed circuit board with six APA102-2020 RGB LEDs, two per color, compared to eight larger LEDs on a commercial board in the original sensor. The LEDs are arranged in an alternating pattern around the camera and covered with a 3D-printed collimator ring that channels the light to create a symmetric illumination pattern on the inside of the sensor (\figref{fig:overview}A). Following Sun et al. \cite{Sun2022:Insight}, we jointly tune the opening angle (20$^\circ$) and diameter (\SI{2.4}{mm}) of the collimator to create a light pattern that fully covers the internal sensing surface.  This design creates both colored structured light and photometric stereo effects, a combination that has been shown by Sun et al.\ to increase the force-sensing accuracy \cite{Sun2022:Insight}.

\subsection{Mechanical Design}

To fit the sensor on humanoid robotic fingers, we reduce Insight's height from 70\,mm to 30\,mm and diameter from 40\,mm to 22\,mm so that the new form factor is comparable to the distal phalanx of a human thumb, as shown in \twofigSref{fig:overview}{B}{fig:minsight_package}{A}.
The silicone shell is over-molded on a stiff metal skeleton to allow for high interaction forces and maintain the sensor's shape. The outer surface has high softness and high friction to facilitate manipulation.

\begin{figure*}[ht!]
    \centering
    \includegraphics[width=1.0\textwidth]{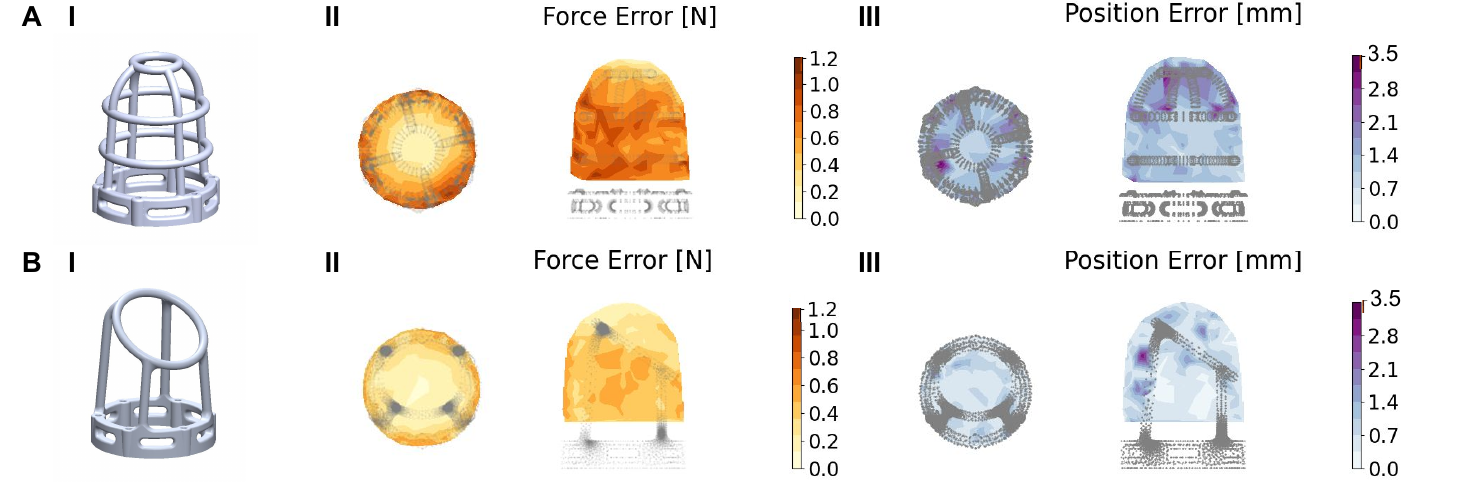}
    \caption{Comparison of two skeleton designs: \textbf{A} has a homogeneous and dense beam structure, whereas \textbf{B} has sparse beams. Both localization and force error are significantly higher on A than on B.}
    \label{fig:skeletons}
\end{figure*}

For the miniaturization, we investigate two different skeleton designs by examining the final prediction errors each can achieve for localization of contact and estimation of force components. 
\figSref{fig:skeletons}{A} shows one possible skeleton structure, with the design objective to create a homogeneous sensor surface while still leaving open spaces for the elastomer to deform. 
\figref{fig:skeletons} B shows a skeleton that is designed to reduce the number of beams to obtain high sensitivity while still being able to withstand high-impact forces. This design leads to larger surface deformations between the beams and makes the surface hardness less homogeneous. The opening at the top of this skeleton is designed to be inclined, similar to the bone inside a human fingertip \cite{serhat_free_2021}. As the sensor is intended for articulated robotic fingers with an anthropomorphic design, this area of the fingertip will experience contact most often. The inclined opening ensures high sensitivity in this area without an obstructing skeleton, as seen in the error plot in \figref{fig:skeletons} B.
The full sensor built with skeleton B shows lower errors for the prediction of the force magnitude as well as lower errors for the prediction of the position of contact, as quantified in \figref{fig:skeletons}. The stiff skeleton reduces the actual displacement of the elastomer surrounding it, which makes changes due to contact forces less discriminative in the camera image, leading to a smaller sensitivity and therefore a larger prediction error in the regions around the skeleton beams.
Thus, skeletons similar to design A can be used for applications with higher contact forces, but skeletons like design B are preferred for more sensitive applications, as selected for the remainder of this article.

\subsection{Manufacturing}

As depicted in \figref{fig:overview} B, the sensor's camera mount and LED collimator are fabricated using a Formlab3 stereo-lithography printer. The stiff skeleton is 3D printed from aluminum alloy to increase strength while maintaining low weight. The elastic shell is over-molded onto the skeleton using a 3D-printed mold.

Compared to the Insight sensor, \sname{}'s smaller size greatly reduces the distance between the camera and the interior surface, therefore requiring a new mixing ratio for the elastomer components. To obtain optimal reflective properties and opacity, we combine \SI{15}{\gram} Smooth-On EcoFlex part A, \SI{15}{\gram} Smooth-On EcoFlex partB, \SI{3}{\gram} aluminum powder, and \SI{0.1}{\gram} aluminum flakes. To ensure homogeneity of the material, we degas the silicone mixture for \SI{5}{\minute} at \SI{5}{\Pa} after mixing and a second time after pouring it into the mold. The over-molded skeleton is then mounted on the connector with six M1.7 screws.

\subsection{Data Processing}

Following \cite{Sun2022:Insight}, we employ a data-driven method to estimate the force distribution directly from the raw input image using deep networks from computer vision (\figSref{fig:overview}{B}). To collect training data for the neural network, we use a position-controlled five-degree-of-freedom (5-DoF) test bed with a spherical indenter that probes the sensor at many locations. A 6-DoF force-torque sensor (ATI Mini40) measures the ground-truth force vector applied to the indenter while the contact location and the camera image from inside the sensor are simultaneously recorded (see \figSref{fig:minsight_package}{C}). The test bed has an overall positioning precision of \SI{0.2}{mm}, and the force-torque sensor has a force-measuring precision of \SIlist{0.01;0.01;0.02}{N} in the $x$-, $y$-, and $z$-directions respectively.

We evaluate the sensor's performance in two modes. The first is single-contact estimation, in which the contact location and the contact force vector are directly inferred. 
The training data for this mode are obtained during the described data collection without further processing.
The second mode is force-distribution estimation: here, we infer the distribution of 3D contact force vectors at an array of points spread across the entire sensing surface. Locations with non-zero force represent the contact area.
The target force-distribution map for training this deep network cannot be measured directly and is thus approximated around the contact point of the indenter, as proposed by Sun et al.~\cite{Sun2022:Insight}. 

Deep neural networks are trained to learn the mapping from the image input to the two types of force output. To highlight changes, a reference image without any contact is subtracted from the RGB image taken at the time of contact. For the force-distribution estimation, the resulting RGB image is also concatenated with a single-channel image that encodes the pixel position before input to that network. The ground-truth measurements described above (force vector and contact location; force vector distribution) are then normalized and used as labels for the supervised training of the mapping networks.

\section{Optimizing the Inference Method}
The inference procedure requires computational resources and introduces a delay that should be reduced as much as possible for practical applications. The minimum goal for real-time performance is to calculate the force-distribution map before a new image is acquired.
We therefore analyze several key parts of the processing pipeline.

\subsection{Force Mapping}
\begin{figure}[tbh]
\includegraphics[width=1.0\textwidth]{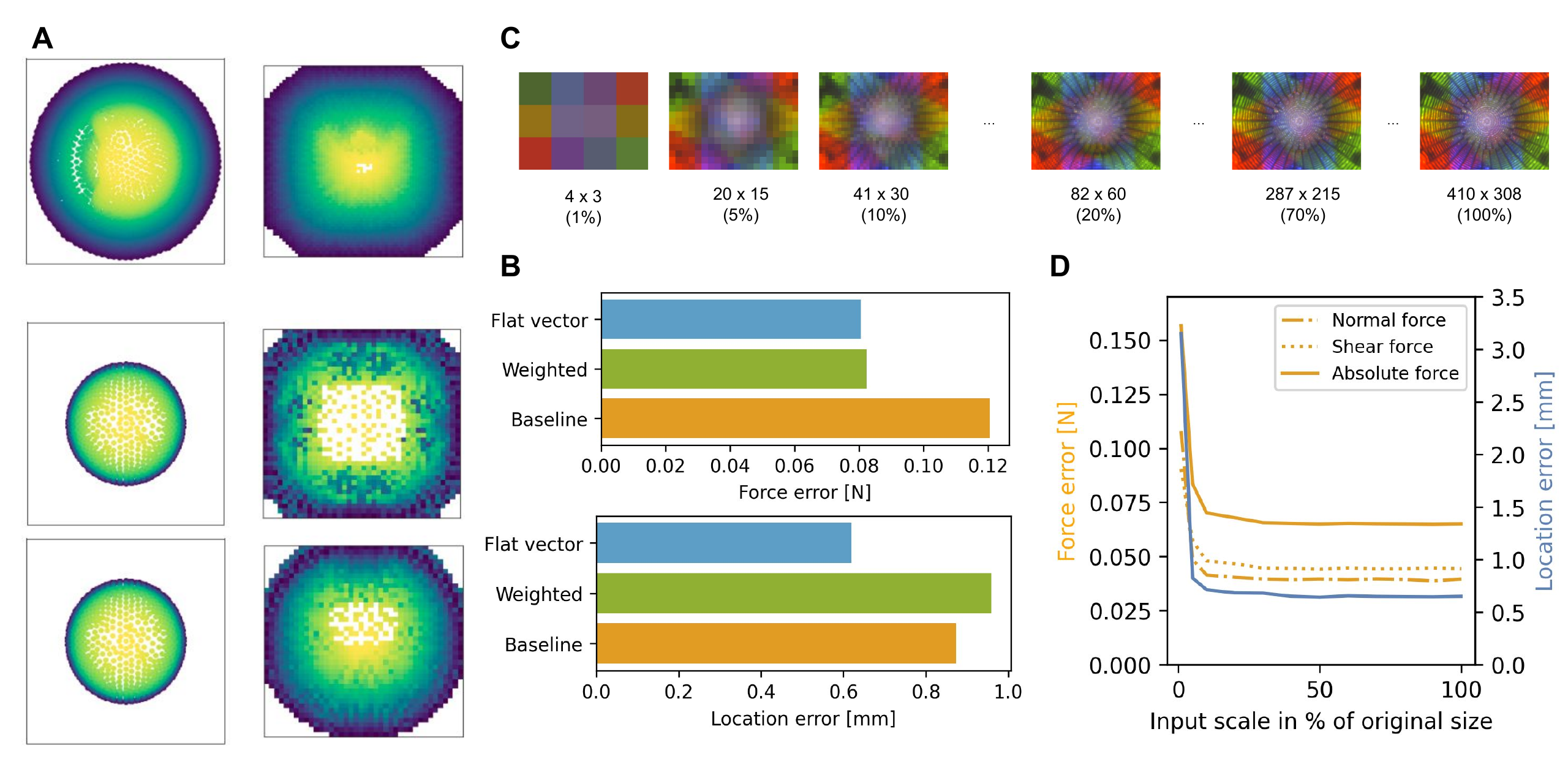}
\caption{Optimization of the processing pipeline that infers haptic information from the camera image. \textbf{A}:~Top view of the distribution of discretized nodes color coded according to their $z$ coordinates (left) and their corresponding distribution in the 2D pixel map created by Hungarian matching (right); note that white pixels in the force map have no associated point in 3D. Top row: results of Hungarian matching for Insight \cite{Sun2022:Insight}, middle row: Hungarian matching for \sname{}, bottom row: weighted Hungarian matching for \sname{}.   \textbf{B}:~Force and location errors for different network output methods. \textbf{C}:~Sample input image at a range of resolutions. 100\% corresponds to the resolution used in Insight. \textbf{D}:~Evaluation of the inference performance on direct force prediction with different input-image resolutions.}
\label{fig:ablation_study}
\end{figure}

To represent the distribution of contact force vectors across \sname{}'s whole sensing surface, we discretize its 3D curved surface with \num{1350} points that are spaced roughly \SI{1}{mm} apart. The machine-learning pipeline is trained to approximate the force-distribution map (represented by a 3-channel image) by using convolutional layers. A mapping from the three-dimensional space to a two-dimensional image is therefore required, which Sun et al.~\cite{Sun2022:Insight} created using the Hungarian assignment method~\cite{kuhn_hungarian_1955}. As shown in \figSref{fig:ablation_study}A, this approach works nicely for the evenly spaced nodes on the cone-shaped Insight sensor. For a sensor with a more cylindrical shape, the assignment method creates a less-smooth and not-neighborhood-preserving mapping, as the density of points on the outer perimeter is much higher than in the center (see \figSref{fig:ablation_study}A, middle row). Since the forces experienced by neighboring points are often correlated, we expect smooth maps to perform better. To mitigate the undesired roughness in \sname{}'s initial map, we shift the zero-centered $x$- and $y$-coordinates of each point by an amount proportional to the ratio between their $z$-coordinate and the maximum height of the sensor, as follows:
\begin{equation}
\label{hungarian_weighted}
x^* = x - \alpha\left(\frac{z}{z_{\max}}\right)x,\hspace{1em} 
y^* = y - \alpha\left(\frac{z}{z_{\max}}\right)y.
\end{equation}
We found that a tiny shifting weight ($\alpha = 0.0005$) is sufficient to draw the sensor's upper points closer to the center of the mapping. The resulting force map has much higher local smoothness, as shown in \figSref{fig:ablation_study}A, bottom row.

A more general mapping from sensor surface points to network output would be desirable for adapting this approach to arbitrary sensor surfaces. We therefore also test the network performance when it is trained to output a flattened vector of all surface points. In the case of \sname{}, this output corresponds to a vector of length \res{1350}{3}, with \num{1350} surface points times \num{3} force vector components. Interestingly, learning to output a flat vector performs just as well as or even better than the originally proposed 2D image, as shown by the force and location errors plotted in \figSref{fig:ablation_study}B. This simpler network output makes it easier to apply this method to different sensor surfaces, as no hand-crafted mapping is necessary. Note, however, that the reported results involve only single contacts. Our preliminary testing indicates that the spatial output representation may outperform the flat vector when the sensor experiences multiple contacts; more investigation is needed to understand these tradeoffs. Thus, we use the weighted spatial output mapping for the remainder of this article.

\subsection{Image Resolution}
\label{sec:resolution}
A significant portion of this system's computational complexity can be traced back to performing inference on high-resolution images. Thus, we consider the influence of input image resolution on the precision of single force and contact location prediction, starting with the already horizontally cropped and down-sampled image size of \res{410}{308}, as used in Insight~\cite{Sun2022:Insight}. \FigSref{fig:ablation_study}C presents a sample \sname{} image with several dimensions down to only \res{4}{3} (\num{1}\% scale). \FigSref{fig:ablation_study}D shows how force error and location error change with input image resolution when employing a ResNet network for inference.
Importantly, a resolution of \res{82}{60} (\num{20}\% scale) is enough for the sensor to predict force with a mean absolute error of \SI{0.07}{N} and location with a mean absolute error of \SI{0.6}{mm}, which are only negligibly higher than the errors at \num{100}\% scale. The results of this analysis indicate that the spatial resolution of the sensor is a function of the entire system and does not solely depend on the camera image resolution. Observing a sensing surface of \SI{1740}{\milli\metre\squared} with an image of \res{82}{60} pixels allows a viewing area of \SI{0.27}{\milli\metre\squared} per pixel, which corresponds to a square patch with an edge length of \SI{0.59}{mm}, matching the spatial resolution of the sensor almost perfectly.  
Compared to Insight, down-scaling the acquired images to \res{82}{60} reduces the number of pixels that must be processed by a factor of \num{25}, enabling a significant speedup with no reduction in accuracy.

\begin{table*}[b!]
\centering
\vspace{0.2cm}
\caption{Errors and inference times of different networks for direct force prediction; GPU: NVIDIA Quadro RTX 6000 graphics card, CPU: Intel Xeon E5-4620 @\SI{2.2}{GHz} on 
\num{1} core, Raspberry Pi: 4B.}
\resizebox{\textwidth}{!}{
\begin{tabular}{ l|r|cc|ccc } 
 \toprule
 \multirow{2}{*}{Network Architecture} &  \multirow{2}{*}{Parameters} & \multicolumn{2}{c|}{Errors} &
 \multicolumn{3}{c}{Mean Inference Time $\pm$ Std.\ Dev.\ [ms]}\\
  & & Force [N] & Location [mm] & GPU & CPU & Raspberry Pi\\
 \midrule
 ConvMixer  \cite{resnet_patches_2022} & 127,806 &0.08 & 1.26 & \textbf{1.41$\pm$0.23} & \textbf{2.39$\pm$0.23} & 45.9$\pm$3.80 \\
  SqueezeNet \cite{iandola_squeezenet_2016} & 725,574 &0.10 & 1.28 & 1.79$\pm$0.15 & 5.9$\pm$0.83 &\textbf{14.5$\pm$2.20} \\
 MobileNet \cite{sandler_mobilenetv2_2019} & 2,231,558 &0.08 & 0.65 & 3.69$\pm$0.33 & 10.29$\pm$0.67 & 334.7$\pm$188.8\\
  EfficientNet \cite{tan_efficientnet_2020} & 4,015,234 &0.14 & 1.25 & 8.69$\pm$0.69 & 15.83$\pm$0.75 &376.8$\pm$104.4\\ 
 ResNet \cite{he_deep_2015} & 11,179,590 &\textbf{0.06} &\textbf{0.52} & 2.01$\pm$0.28 & 29.42$\pm$1.14 &75.2$\pm$6.10  \\
 \bottomrule
\end{tabular}}
\label{tab:network_errors_and_latency}
\end{table*}

\subsection{Network Architecture}
\label{sec:network}
To further decrease the need for computational power, we train the task of direct force prediction on four state-of-the-art image-processing networks that feature fewer parameters than the ResNet18 used in the original work~\cite{Sun2022:Insight}, comparing their prediction accuracy and inference speed. For each selected network architecture, \tabref{tab:network_errors_and_latency} shows the number of parameters it involves, the prediction errors it achieves for contact force and contact location error, and its mean inference time on an NVIDIA Quadro RTX6000 GPU. Although ResNet18 delivers the lowest errors, other network architectures with far fewer parameters can deliver similar sensing performance. A SqueezeNet implementation of the processing pipeline can even keep up with the camera frame rate of \SI{60}{Hz} when deployed on a Raspberry Pi 4B with \num{2} GB RAM. Such fast inference calculations alleviate the need for powerful compute hardware for data processing and make the proposed sensor suitable for deployment on mobile robots.

\section{Applications}
\label{application}
To further validate the sensor's performance and test whether it can provide useful tactile information, we mount it at the end of one 3-DoF manipulator of the TriFinger robot platform \cite{wuthrich_trifinger_2020}. We wrap the sensor processing pipeline in a ROS node and integrate it into the real-time control loop of the chosen manipulator. Two tactile applications are explored with this setup: tactile servoing and lump detection. Videos of both can be found in the supplementary material. 

 \begin{figure*}[b!]
    \includegraphics[width=\linewidth]{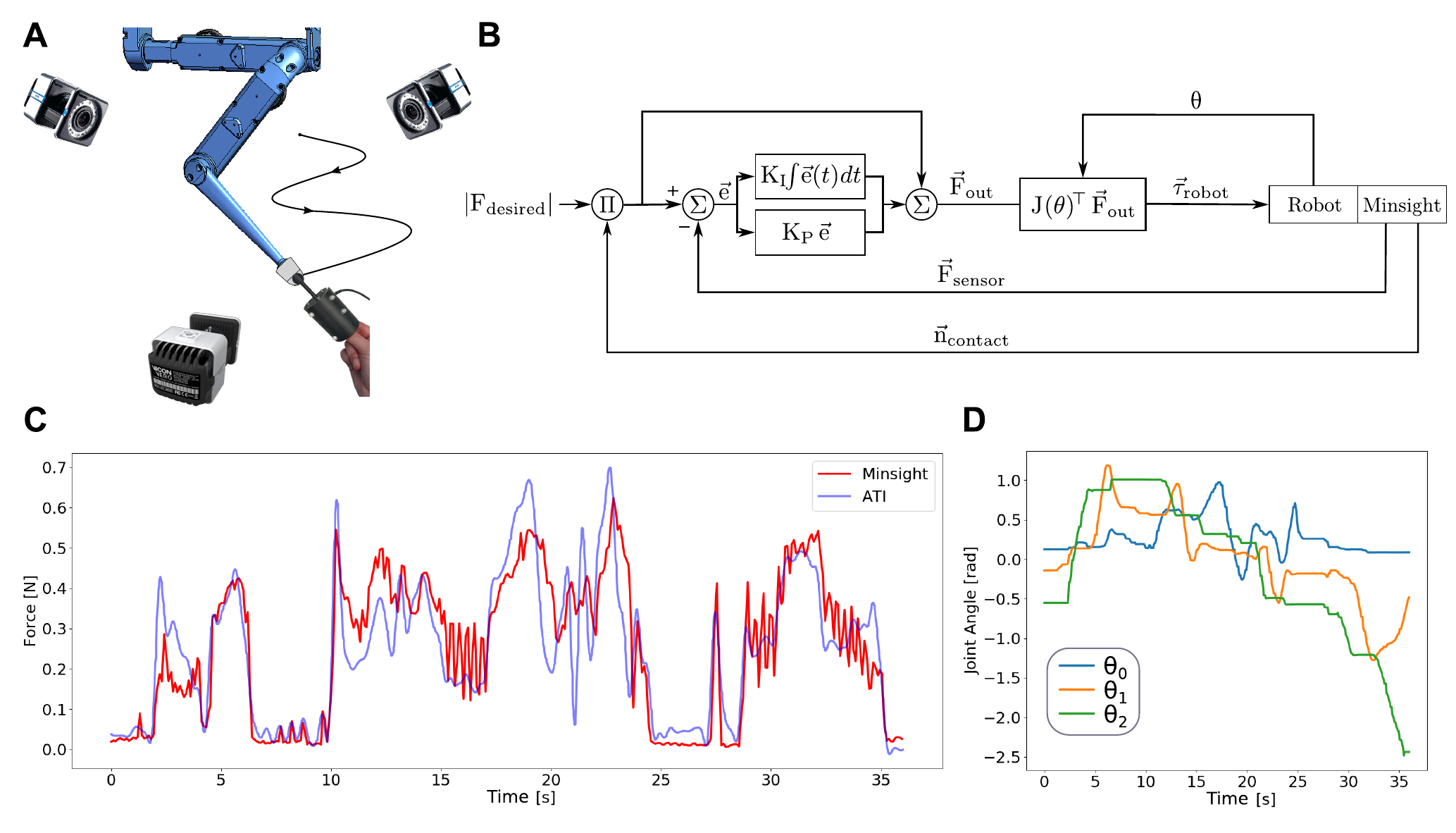}
    \caption{The tactile servoing experiment. \textbf{A}:~Experiment setup: The user contacts \sname{} with an indenter mounted on a three-axis force sensor. The force-torque sensor itself is placed inside a 3D-printed housing with motion-tracking markers. Four cameras track the orientation of the measurement device. \textbf{B}:~The feedback control loop used to achieve force control with the depicted setup. 
    The forces in end-effector space $\vec F_\mathrm{out}$ are regulated by a proportional and integral (PI) feedback controller and are transformed into joint space using the transpose of the robot's Jacobian ($J(\theta)^\top$). 
    \textbf{C}:~\sname{} output data compared to the low-pass-filtered ground-truth force measurements. \textbf{D}:~Joint angles of the robot recorded during the tactile servoing experiment}\label{fig:tactile_servoing}
 \end{figure*}
\subsection{Tactile Servoing}
In the first experiment, the sensor output is used to control the manipulator in a force-controlled fashion to follow a user's finger by trying to maintain a specified contact force within a preset range along the sensed contact normal (\figSref{fig:tactile_servoing}A). The controller includes a feed-forward term as well as proportional and integral (PI) feedback, as shown in \figSref{fig:tactile_servoing}B. 
To collect ground-truth force-vector data, we use an ATI Mini40 force-torque sensor mounted in a handheld case to measure three-axis contact forces between the 16\,mm indenter and the sensor. The Mini40's case is equipped with markers that are tracked by a four-camera Vicon motion-capture system installed around the robot. We use the orientation of the force sensor relative to the global coordinate system to calculate the ground-truth contact force in a common world coordinate frame, subtracting the weight of the indenter. Using the robot joint angles, we also transform the \sname{} sensor readings into the world coordinate frame so that they can be compared along each axis. Force measurements from the ATI sensor, the tracked orientation of the ATI case, and the robot joint angles were recorded at \SI{100}{Hz}. \FigSref{fig:tactile_servoing}{D,C} show the robot's joint angles for one sample trial, along with the recorded contact forces. To reduce noise, the ground-truth contact force measurements were post-processed with a low-pass filter with a cutoff frequency of \SI{30}{Hz}. The forces estimated by \sname{} closely match those measured with the ground-truth force sensor, and the robot easily tracks the finger of the user, proving its suitability for real-time robot control applications.

\subsection{Lump Classification}
\begin{figure*}[t!]
    \includegraphics[width=\linewidth]{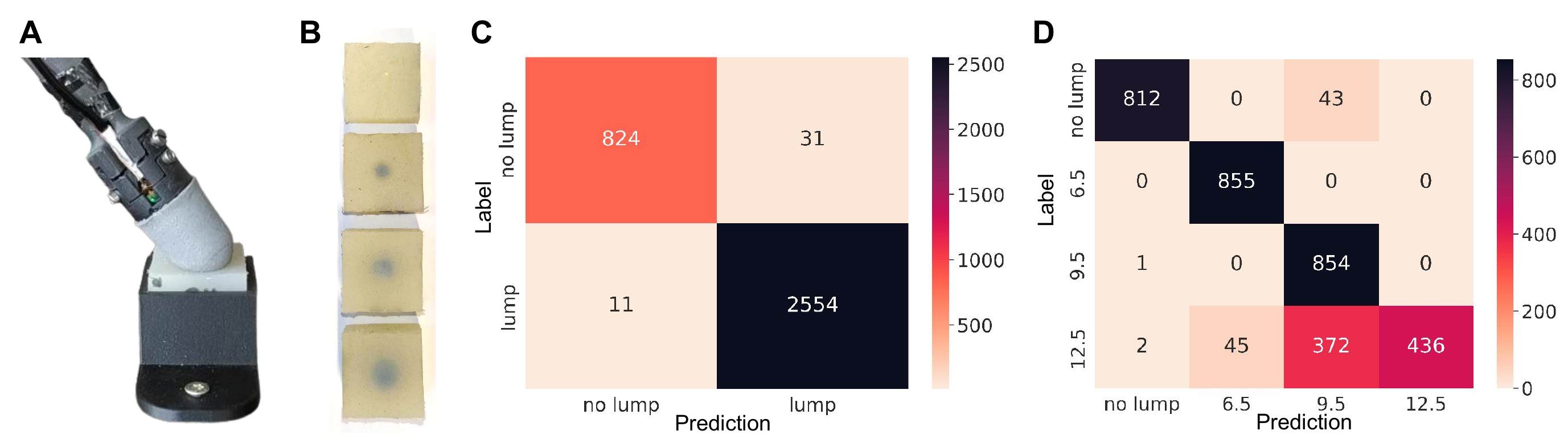}
    \caption{\textbf{A}: The manipulator is commanded to press into the soft material at random positions while \sname{}'s output is continually recorded. \textbf{B}: Four different classes are used: no lump, \SI{6.5}{mm} lump, \SI{9.5}{mm} lump, and \SI{12.5}{mm} lump. 
    \textbf{C}: Confusion matrix on a holdout test set for a binary classification between samples without and with lumps. \textbf{D}: Confusion matrix on a holdout test set for multi-class classification of different lump sizes. }
    \label{fig:lump_detection}
 \end{figure*}

Tactile sensing is required to detect differences in hardness within an opaque soft object similar to human tissue; this clinically motivated palpation task cannot be solved with conventional vision-based approaches. To showcase the capabilities of \sname{}, we conduct lump classification with eight different rectangular samples made of EcoFlex30. Six samples contain an embedded hard sphere with a diameter of \num{6.5}, \num{9.5}, or \SI{12.5}{mm} (two samples for each size), and the other two contain no sphere (see \figSref{fig:lump_detection}B). We command the \sname{}-equipped 3-DoF manipulator to press down in a position-controlled fashion at random positions on top of the soft sample and record the sensor output (see \figref{fig:lump_detection}). We then train a small neural network to do classification on single force map outputs at every time step along the robot trajectory. For this purpose, we split all recorded trajectories into training and testing sets, to guarantee that the network has not seen any part of the testing trajectory during training. To ensure that contact between the sensor and the sample has been made, we discard the first few samples of each trajectory and classify only force maps with a total estimated contact force higher than \SI{0.3}{N}. The classification results in \figSref{fig:lump_detection}C show an accuracy of \num{98}\% on a binary classification task between samples without an embedded lump and all samples with lumps. From \figref{fig:lump_detection} D we can see that classification between samples without a lump and lumps of sizes \num{6.5} and \SI{9.5}{mm} is very accurate, while samples with an embedded \SI{12.5}{mm} lump are sometimes confused with smaller lumps. The overall accuracy for multi-class lump classification is \num{89}\%, which is \num{3.56} times better than chance. We hypothesize that the largest lump was most challenging to identify because its tactile impression feels similar to the smaller lumps at low levels of pressing force.

\section{Conclusion}
\label{sec:conclusion}
This paper presents a fingertip-sized haptic sensor that accurately estimates distributed normal and shear forces, as well as contact location. \sname{} is very sensitive but still inherently robust and low-cost, also featuring favorable friction properties for handling objects. The real-time capabilities of the sensor are demonstrated through two robotic applications that successfully showcase its robustness and reliability.
As shown in Sections \ref{sec:resolution} and \ref{sec:network}, the spatial resolution of the sensor is not limited by the input image resolution, the test bed accuracy or the network capacity. More likely, the physical smoothing properties of the elastomer and the shell design are the limiting factors. The current design has a spatial resolution of \SI{0.6}{mm}, which is already lower than the tactile spatial resolution of humans, which was experimentally determined to be \SI{0.92}{mm} \cite{VanBoven2361}.
The sensor can stream data at a maximum frequency of \SI{60}{Hz}, limited only by the frame rate of the chosen camera. A higher sampling rate that is closer to the human tactile sensing bandwidth would be possible with the development of even faster small-scale cameras. We are also looking into integrating vibration sensing so that the sensor could be more sensitive to fine textures and transient effects such as slip. Another promising future direction is the incorporation of mechanoluminescent or electroluminescent materials \cite{PENG202191} into the sensor shell, which could complement or eliminate the need for lighting within the sensor. The use of self-healing materials in optical sensors, as proposed by Bai et al. \cite{selfhealing_shepherd}, is another exciting direction to extend the capabilities of tactile sensors toward human performance. Nonetheless, we believe that the presented approach to fingertip tactile sensing can already equip robot manipulators with a good sense of touch for dexterous manipulation and physical human-robot interaction. Furthermore, \sname{}'s simple, low-cost design will facilitate further exploration of practical tactile sensing for robotic applications.

\subsection*{Supporting Information}  %
The data that support the findings of this study are openly available in Edmond at \url{https://doi.org/10.17617/3.AEDHD1} and code to reproduce the reported results can be found on GitHub at \url{https://github.com/martius-lab/Minsight-sensor}.

\subsection*{Author Contributions}
I.A., H.S., K.J.K., and G.M. conceived the method and the experiments.
H.S. initiated the project, customized the camera, and then handed over project leadership to I.A.
I.A. designed and constructed the sensor hardware with hands-on guidance from H.S. 
I.A. designed and conducted experiments for optimization of the inference method, collected data, and analyzed the data, all with guidance from H.S.
I.A. conducted both robotic application experiments, including collecting and analyzing the data, with supervision from K.J.K. and G.M.
I.A. drafted the manuscript, and all authors revised it.

\subsection*{Acknowledgements}
The authors thank the International Max Planck Research School for Intelligent Systems (IMPRS-IS) for supporting Iris Andrussow and Huanbo Sun. Furthermore, they thank Julian Martus and Jonathan Fiene for helping with the design and manufacturing of the LED ring, as well as Felix Widmaier for helping with the interface to the robotic manipulator.

\medskip

\printbibliography

\end{document}

%% file: header.tex
\usepackage{times}
\usepackage[utf8]{inputenc} %
\usepackage[T1]{fontenc}    %
\usepackage{url}            %
\usepackage{booktabs}       %
\usepackage{nicefrac}       %
\usepackage{microtype}      %
\usepackage{graphics,color}

\usepackage{algorithm}
\usepackage{algorithmic}
\usepackage{enumitem}

\usepackage{amsfonts}       %
\usepackage{amsmath}       %
\usepackage{amssymb}

\usepackage[doi=false,isbn=false,url=false,eprint=false,natbib=true,style=numeric,sorting=none, maxbibnames=99,giveninits=true]{biblatex} %

\usepackage{subcaption}

\usepackage{etoolbox}
\robustify\,

\usepackage{siunitx}
\usepackage{multirow}
\usepackage{stfloats}
\usepackage{float}
\usepackage{parskip}
\usepackage{svg}

\newcommand{\FigSref}[2]{Figure~\ref{#1}\,#2}  %
\newcommand{\figref}[1]{Fig.~\ref{#1}}    %
\newcommand{\figSref}[2]{Fig.~\ref{#1}\,#2}    %
\newcommand{\twofigSref}[4]{Figs.~\ref{#1}\,#2 and \ref{#3}\,#4} %

\newcommand{\tabref}[1]{Table~\ref{#1}}

\newcommand{\secref}[1]{Sec.~\ref{#1}} %

\usepackage{xspace}
\makeatletter
\DeclareRobustCommand\onedot{\futurelet\@let@token\@onedot}
\def\@onedot{\ifx\@let@token.\else.\null\fi\xspace}
\makeatother

\usepackage{xr-hyper}
\makeatletter
\newcommand*{\addFileDependency}[1]{%
  \typeout{(#1)}
  \@addtofilelist{#1}
  \IfFileExists{#1}{}{\typeout{No file #1.}}
}
\makeatother

\usepackage{xcolor}
\definecolor{ourblue}{rgb}{0.368,0.507,0.71}
\definecolor{ourorange}{rgb}{0.881,0.611,0.142}
\definecolor{ourgreen}{rgb}{0.56,0.692,0.195}
\definecolor{ourred}{rgb}{0.923,0.386,0.209}
\definecolor{ourviolet}{rgb}{0.528,0.471,0.701}
\definecolor{ourbrown}{rgb}{0.772,0.432,0.102}
\definecolor{ourlightblue}{rgb}{0.364,0.619,0.782}
\definecolor{ourdarkgreen}{rgb}{0.572,0.586,0.}

\definecolor{ourcyan2}{rgb}{0.125,0.722,0.804}
\definecolor{ourred2}{rgb}{0.863,0.184,0.047}
\definecolor{ouryellow2}{cmyk}{0,0.16,1.0,0.07}
\definecolor{ourviolet2}{cmyk}{0.55,0.56,0,0.47}
\definecolor{ourorange2}{cmyk}{0,0.46,0.89,0.11}

\usepackage{makecell}

%% file: commands.tex
\newcommand{\sname}{Minsight\xspace}
\newcommand{\res}[2]{\ensuremath{#1\!\times\!#2}}